\documentclass[10pt,twocolumn,letterpaper]{article}

\usepackage{cvpr}
\usepackage{times}
\usepackage{epsfig}
\usepackage{comment}
\usepackage{times}
\usepackage{graphicx}
\usepackage{lineno}
\usepackage{color}
\usepackage{algorithm}
\usepackage{algorithmic}
\usepackage{subfigure}
\usepackage[fleqn]{amsmath}
\usepackage{amssymb}
\usepackage{fancyhdr}
\renewcommand{\headrulewidth}{0pt}

\def\Tr{{\mbox{Tr}}}

\def\AAT{{$P \kern-2.0pt P \kern-1.0pt ^T$ }}
\def\ATA{{$P \kern-1.0pt ^T \kern-2.0pt P$ }}
\def\AATe{{X \kern-2.0pt X \kern-1.0pt ^T }}
\def\ATAe{{X \kern-1.0pt ^T \kern-2.0pt X }}

\def\ub{{\bf u}}

\def\vb{{\bf v}}

\def\pdd{{p_{\kern-2pt\lower+1pt\hbox{..}}\kern+1pt }}
\def\pdds{{p_{\kern-2pt\lower+1pt\hbox{..}}^{1/2}\kern+1pt}}
\def\sdd{{s_{\kern-2pt\lower+1pt\hbox{..}}\kern+1pt }}
\def\sdds{{s_{\kern-2pt\lower+1pt\hbox{..}}^{1/2}\kern+1pt}}

\def\pdc#1{{p_{\kern-2pt\lower+1pt\hbox{.}\lower+2pt\hbox{\footnotesize $#1$}\kern+1pt }}}
\def\sdc#1{{s_{\kern-2pt\lower+1pt\hbox{.}\lower+2pt\hbox{\footnotesize $#1$}\kern+1pt }}}

\def\prd#1{{p_{\kern-1pt\lower+2pt\hbox{\footnotesize $#1$}\lower+1pt\hbox{.}\kern+1pt }}}
\def\srd#1{{s_{\kern-1pt\lower+2pt\hbox{\footnotesize $#1$}\lower+1pt\hbox{.}\kern+1pt }}}

\def\rb{{\bf r}}

\def\wb{{\bf w}}

\def\vv1k{{{\bf v}^1 \cdots {\bf v}^k}}
\def\xxx1n{{{\bf x}_1 \cdots {\bf x}_n}}

\def\Dm1{{D^{-1}}}
\def\Dm12{{D^{-1/2}}}
\def\Dp12{{D^{1/2}}}
\def\Dxm12{{D_r^{-1/2}}}
\def\Dym12{{D_c^{-1/2}}}
\def\Dxp12{{D_r^{1/2}}}
\def\Dyp12{{D_c^{1/2}}}

\def\YYT{{Y \kern-0.0pt Y \kern-1.0pt ^T }}
\def\YTY{{Y \kern-1.0pt ^T \kern-0.0pt Y }}
\def\YT{{Y \kern-1.0pt ^T}}
\def\XT{{X \kern-1.0pt ^T}}
\def\XTX{{X \kern-1.0pt ^T \kern-2.0pt X }}

\def\FFTD {{F \kern-1.0pt F^T \kern-3.0pt D }}

\def\bbox{{\hfill {$\sqcap \kern-6.0pt \lower+2.4pt\hbox{--} \kern+2.7pt $}}}
\def\squa {{{$\sqcap \kern-6.0pt \lower+2.4pt\hbox{--} \kern+2.7pt $}}}

\def\Jpara{J_{\mbox{\scriptsize ParaFac}}}
\def\Jcf{J_{\mbox{\tiny T1}}}

\usepackage[pagebackref=true,breaklinks=true,letterpaper=true,colorlinks,bookmarks=false]{hyperref}

\cvprfinalcopy

\def\cvprPaperID{1410} 

\ifcvprfinal\fi
\begin{document}
\title{Are Tensor Decomposition Solutions Unique?\\
On the global convergence of HOSVD and ParaFac algorithms}

\author{Dijun Luo, Heng Huang, Chris Ding\\
Computer Science and Engineering\\
University of Texas at Arlington\\
Technical Report CSE-2009-5
}
\maketitle
\thispagestyle{fancy}

\begin{abstract}
 For tensor decompositions such as HOSVD and ParaFac, the
 objective functions are nonconvex.
 This implies, theoretically,
 there exists a large number of local optimas:
 starting from different starting point, the iteratively improved
 solution will converge to different local solutions.
 This non-uniqueness present a stability and reliability problem for image
 compression and retrieval. In this paper,
 we present the results of a comprehensive investigation of this problem.
 We found that although all tensor decomposition algorithms
 fail to reach a unique global solution on random data and severely scrambled
 data; surprisingly however, on all real life several data sets (even with
 substantial scramble and occlusions), HOSVD always
 produce the unique global solution in the parameter region
 suitable to practical applications, while ParaFac produce non-unique
 solutions. We provide an eigenvalue based rule
 for the assessing the solution uniqueness.
\end{abstract}


\section{Introduction}

Tensor based dimension reduction has recently been extensively
studied for computer vision, pattern recognition, and machine
learning applications. Typically, such approaches seek subspaces such
that the information are retained while the discared
subspaces contains noises.
Most tensor decomposition methods are unsupervised which enable
researchers to apply them in any machine learning applications
including unsupervised learning and semi-supervised learning.

Perhaps High Order Singular Value Decomposition (HOSVD)
\cite{Tucker66}~\cite{Lathauwer:HOSVD}
and Parallel Factors (ParaFac)
are some of the most widely used tensor decompositions.
Both of them could be viewed as extensions of
SVD of a 2D matrix.
HOSVD is used in computer vision by
Vasilescu and Terzopoulos~\cite{HOSVD}
while
ParaFac is used in computer vision by
Shashua and Levine \cite{Shashua:cvpr}.
More recently,
Yang {\em et al.} \cite{2DPCA}
proposed a two dimensional PCA (2DPCA)
Ye {\em et al.} \cite{GLRAM} proposed a
method called Generalized Low Rank Approximation of Matrices (GLRAM).
Both GLRAM and 2DPCA can be viewed in the same framework in
2DSVD (two-dimensional singular value decomposition)
\cite{2dsvd}.
and solved by
non-iterative algorithm
\cite{Inoue:2dsvd}
The error bounds of HOSVD have been derived \cite{Ding:HOSVD-error}
and the equivalence between tensor $K$-means clustering and
HOSVD is also established \cite{Huang:kdd08}.

Although tensor decompositions are now widely used, many of their
properties so far have not been well characterized.
For example, the tensor rank problem remains a research issue.
Counter examples exist that argue against optimal low-dimension approximations
of a tensor.

In this paper, we address the solution uniqueness issues.
This problem arises because
the tensor decomposition objective functions are
non-convex with respect to all the variables
and the constraints of the optimization are also non-convex.
Standard algorithms to compute these decompositions are iterative
improvement. The non-convexity of the optimization
implies that the iterated solutions
will converge to different solutions if they start from different
initial points.

Note that this fundamental uniqueness issue differs from other
representation redundancy issues, such as
equivalence transformations (i.e. rational invariance)
that change individual factors $(U,V,W)$
but leaves the reconstructed image untouched.
These representation redundancy issues can be
avoided if we compare different solutions at the
level of reconstructed images, rather in the level
of individual factors.

The main findings of our investigation
are both surprising and comforting.
On all real life datasets we tested
(we tested 6 data sets and show results for
3 data set due to space limitation),
the HOSVD solutions are unique (i.e., different
initial starts always converge to an unique global solution);
while
the ParaFac solution are almost always not unique.
Furthermore, even with substantial randomizations
(block scramble, pixel scramble, occlusion) of
these real datasets,
HOSVD converge to unique solution too.

These new findings
assure us that in most applications using HOSVD, the solutions are
unique --- the results are repeatable and reliable.

We also found that
whether a HOSVD solution is unique can be
reasonably predicted by
inspecting the eigenvalue distributions
of the correlation matrices involved.
Thus the eigenvalue distributions
provide a clue about the solution uniqueness
or global convergence. We are looking
into a theoretical explanation of this rather robust uniquenss
of HOSVD.

\section{Tensor Decomposition}

\subsection{High Order SVD (HOSVD)}
Consider 3D tensor: $ X = \{ {{ X_{ijk}
\}_{i=1}^{n_1} }_{j=1}^{n_2} }_{k=1}^{n_3}. $
The objective of HOSVD
is to select subspace  $U,V,$, $W$ and core tensor
$S$ such that the $L_2$ reconstruction error is minimized,
\begin{equation}
\min_{U, V, W, S} \; J_1= ||X -  U \otimes_1 V \otimes_2 W \otimes_3
S||^2
\label{EQ:JHOSVD1}
\end{equation}
where
$
U \in \Re^{n_1 \times m_1}, \
V \in \Re^{n_2 \times m_2}, \
W \in \Re^{n_3 \times m_3}, \
S \in \Re^{m_1 \times m_2 \times m_3}.
$
Using explicit index,
\begin{equation}
J_1=\sum_{i j k} \Big( X_{ijk} - \sum_{p q r} U_{ip}
V_{jq}  W_{kr}  S_{pqr}\Big)^2 .
\label{EQ:JHOSVD2}
\end{equation}
In HOSVD, $W, U,V$ are required to be orthogonal:
$$
U^T U = I, \ V^T V = I, \ W^T W = I.
$$
With the orthonormality condition,
setting $\partial J_1 / \partial S=0$,
we obtain
$
S =U^T \otimes_1 V^T \otimes_2 W^T \otimes_3 X,
$
and
$ J_1 = \| X\|^2 - \| S\|^2$.
\label{EQ:UVWX}
Thus HOSVD is equivalent to maximize
\begin{eqnarray}
\max_{U,V,W}
\| S\|^2
&=&
\|U^T \otimes_1 V^T \otimes_2 W^T \otimes_3 X\|^2
\\
&=&
\label{EQ:HOSVD-U}
 \Tr \ U^T F U \\
&=&
\label{EQ:HOSVD-V}
 \Tr \ V^T G V \\
&=&
 \Tr \ W^T H W.
\label{EQ:HOSVD-W}
\end{eqnarray}
where
\begin{eqnarray}
F_{i i'}&=& \sum_{j j' \ell \ell'} X_{ij\ell} X_{i' j'
\ell'} (VV^T)_{j j'}  (WW^T)_{\ell \ell'}
\label{EQ:F}
\end{eqnarray}
\begin{eqnarray}
G_{j j'}&=& \sum_{i i' \ell \ell'} X_{ij\ell} X_{i' j'
\ell'} (UU^T)_{i i'}  (WW^T)_{\ell \ell'}
\label{EQ:G}
\end{eqnarray}
\begin{eqnarray}
  H_{\ell \ell'}&=& \sum_{i i' j j'} X_{ij\ell}
    X_{i' j' \ell'}   (UU^T)_{i i'}  (VV^T)_{j j'}
\label{EQ:H}
\end{eqnarray}
Standard HOSVD algorithm starts with initial guess of
of $(U,V,W)$ and
solve Eqs(3,4,5) alternatively using
eigenvectors of the corresponding matrix.
Since $F,G,H$ are semi-positive definite,
$||S||^2$ are
monotonically increase (non-decrease).
Thus the algorithm converges to a {\it local}  optimal solution.

HOSVD is a nonconvex optimization problem:
The objective function of Eq.(2) w.r.t.
$(U,V,W)$ is nonconvex and
the orthonormality constraints of Eq.(2) are nonconvex as well.
It is well-known that for
nonconvex optimization problems, there are
many local optimal solutions:
starting from different initial guess of
$(U,V,W)$, the converged solutions are
different.
Therefore theoretically, solutions of HOSVD
are not unique.

\subsection{ParaFac decomposition}

ParaFac decomposition~\cite{Harshman:70,Carroll:70}
is the simplest and also most widely used decomposition model.
It approximates the tensor as
\begin{equation}
X \approx \sum_{r=1}^R \ub^{(r)} \otimes \rb^{(r)} \otimes  \wb^{(r)}, \mbox{\ or } \;
 X_{ijk} \approx \sum_{r=1}^R U_{ir}  V_{jr}  W_{kr}
\label{EQ:ParaFac1}
\end{equation}
where $R$ is the number of factors
and
$ U= (\ub^{(1)}, \cdots, \ub^{(R)})$,
$ V= (\vb^{(1)}, \cdots, \vb^{(R)})$,
$ W= (\wb^{(1)}, \cdots, \wb^{(R)})$.
ParaFac minimizes the objective
\begin{eqnarray}
\Jpara
= \sum_{i=1}^{n_1}  \sum_{j=1}^{n_2} \sum_{k=1}^{n_3}
||  X_{ijk}   - \sum_{r=1}^R  U_{ir}  V_{jr}  W_{kr} ||^2
\label{EQ:ParaFac2}
\end{eqnarray}
We enforce the implicit constraints that
columns of $U=(u^{(1)},\cdots, u^{(R)})$ are linearly independent;
columns of $V=(v^{(1)},\cdots, v^{(R)})$ are linearly independent;
and
columns of $W=(w^{(1)},\cdots, w^{(R)})$ are linearly independent.

Clearly the ParaFac objective function is
nonconvex in $(U,V,W)$. The
linearly independent constraints are also nonconvex.
Therefore, the ParaFac optimization is a nonconvex optimization.

Many different computational algorithms were developed for computing
ParaFac. One type of algorithm uses a sequence of
rank-1 approximations~\cite{T.Zhang:rank1,Kolda:parafac,Shashua:cvpr}.
However, the solution of this heuristic approach differ from
(local) optimal solutions.

The standard algorithm is to compute one factor at a time in an alternating
fashion. The objective decrease monotonically in each step, and
the iteration converges to a (local) optimal solution.
However, due to the nonconvexity of ParaFac optimization, the converged
solution depends heavily on the initial starting point. For this reason,
the ParaFac is often not unique.

\section{Unique Solution}

In this paper, we investigate the
problem of
whether the solution of a tensor decomposition is unique.
This is an important problem, because if the solutions is
not unique, then the results are not repeatable and the
image retrieval is not reliable.

For a convex optimization problem,
there is only one local optimal solution which
is also the global optimal solution.
For a non-convex optimization problem,
there are many (often infinite) local optimal solutions:
converged solutions of the HOSVD/ParaFac iterations
depend on the initial starting point.

In this paper, we take the experimental approach.
For a tensor decomposition
we run many runs with dramatically different
starting points.
If the solutions of all these runs agree with each other
(to computer machine precision),
then we consider the decomposition has a unique solution.

In the following, we explain the
(1) The dramatically different starting point for $(U, V, W)$.
(2) Experiments on three different real life data sets.
(3) Eigenvalue distributions which can predict the uniques
of the HOSVD.

\section{A natural starting point for $W$: the T1\\
decomposition and the PCA solution}

In this section, we describe a natural starting point for $W$.
Consider the T1 decomposition~\cite{Tucker66}
\begin{eqnarray}
X_{ijk} \approx  \sum_{k' =1}^{m_3} C_{ij k' }  W_{k k' }
\;\; \mbox{\ or  }  \;\;
X^{(k)}_{ij}   \approx
\sum_{k' =1}^{m_3}  C^{(k')}_{ij} W_{k k'}.
\label{EQ:T1-0}
\end{eqnarray}
$C,W$ are obtained as the results of the optimization
\begin{equation}
\min_{C, W} \;
\Jcf= \sum_{k=1}^{n_3} ||X^{(k)} - \sum_{k'=1}^{m_3}  C^{(r)} W_{kk'}||^2.
\label{EQ:J-T1a}
\end{equation}
This decomposition can be reformulated as the following:
\begin{equation}
\Jcf =
||X||^2 -\Tr  \ (W^T \tilde H W ),
\label{EQ:Jcf<Jpara}
\end{equation}
where
\begin{equation}
\tilde H _{k k'}
=\Tr \ ( X^{(k)} [X^{(k')}]^T)
= \sum_{i  j } X_{ij k} X_{i j k'}   .
\label{EQ:H1}
\end{equation}
$C$ is given by
$
C^{(r)} = \sum_{k=1}^{n_3} X^{(k)} W_{kr}.
\label{EQ:Cr}
$

This solution is also the PCA solution. The reason is the
following. Let
$A = (a_1, \cdots, a_{n} )$
be a collection of 1D vectors.
The corresponding covariance matrix is $A A^T$ and
Gram matrix is $A^T A$.
Eigenvectors of  $A^T A$ are the principal components.
Coming back to the T1 decomposition,
$\tilde H$ is the Gram matrix if we consider each image $X^{(k)}$
as a 1D vector. Solution for $W$ are principal
eigenvectors of  $\tilde H$, which are the principal components.

\section{Initialization}

For both HOSVD and ParaFac, we generate 7 different initializations:

(R1) Use the PCA results $W$ as explained in \S 4. Set $V$ to identity
   matrix (fill zeros in the rest of the matrix to fit
   the size of $n_2\times m_2$). This is our standard initialization.

(R2) Generate 3 full-rank matrixes $W$ and $V$ with uniform random
numbers of in $(0,1)$.

(R3) Randomly generate 3 rank deficient matrices $W$ and $V$
with proper size.
For first initialization, we randomly pick a column
of $W$ and set the column to zero.
The rest of columns are randomly generated as in (R2) and the same for $V$.
For second and third initializations, we randomly pick two
or three columns of $W$ and set them to zero, and so on.
Typically, we use $m_1=m_2=m_3=5\simeq 10$.
Thus the rank-deficiency at $m_3=5$ is strong.

We use the tensor toolbox~\cite{Kolda:toolbox}.
The order of update in
the alternating updating algorithm is the
following:
(1) Given $(V,W)$, solve for $U$ (to solve Problem \ref{EQ:HOSVD-U});
(2) Given $(U,W)$, solve for $V$ (Problem \ref{EQ:HOSVD-V});
(3) Given $(U,V)$, solve for $W$ (Problem \ref{EQ:HOSVD-W});
Go back to (1) and so on.


\section{Run statistics and validation}

For each dataset with each parameter setting, we run 10
independent tests. For each test,
we run HOSVD iterations to convergences
(because of the difficulty of estimating convergence criterion,
we run total of T=100 iterations of alternating updating which
is usually sufficient to converge).

For each independent test, we have 7 different solutions of
$(U_i,V_i,W_i)$ where $i=1, 2, \cdots, 7$ for
the solution starting from the $i$-th initialization.
We use the following difference to verify whether the
solutions are unique:
\begin{equation}
\nonumber
d(t) = \frac{1}{6}\sum_{i=2}^7 \left(\|U_i^t-U_1^t\| +
\|V_i^t-V_1^t\| + \|W_i^t-W_1^t\| \right),
\end{equation}
where we introduce the HOSVD iteration  index $t$, and
$U_i^t, V_i^t, W_i^t$ are the solution in $t$-th iteration.

If an optimization problem has a unique solution,
$d(t)$ typically starts with nonzero value
and gradually decrease to zero.
Indeed, this occurs often in Figure 2
The sooner $d(t)$ decreases to zero, the faster the algorithm converges.
For example, in the 7th row of Figure 2,
the $m_1=m_2=m_3=5$ parameter setting, the algorithm converges faster than
the $m_1=m_2=m_3=10$ setting.

In our experiments, we do 10 different tests (each with
different random starts).
If in all 10 tests $d(t)$ decreases to zero,
we say the optimization has a unique solution
(we say they are globally convergent).

If an optimization has no unique solution (i.e., it has many local optima),
$d(t)$ typically remains nonzero at all times, we say the solution
of HOSVD is not unique.
In Figure~\ref{fig:random},
we show the results of HOSVD and ParaFac on a random tensor.
One can see that in each of the 10 tests,
shown as 10 lines in the figure, none of them ever decrease to zero.

For ParaFac we use the difference of
reconstructed tensor to evaluate the uniqueness of the solution:
\begin{equation}
d'(t) = \frac{1}{6}\sum_{i=2}^7 \|\widehat X_i^t-\widehat X_1^t\| ,
\end{equation}
where $\widehat X_i^t$ is the reconstruction tensor in the $t$-th iteration
with the $i$-th starting point.
ParaFac algorithm converge slower than HOSVD algorithm.
Thus we run 2000 iterations for each test.


\section{Eigenvalue Distributions}

In these figures, the eigenvalues of $F$, $G$, and $H$ are
calculated using Eqs.(\ref{EQ:F},\ref{EQ:G},\ref{EQ:H}),
but setting all $UU^T, VV^T, WW^T$ as identity matrix.
The matrices are centered in all indexes.
The eigenvalues are sorted and
normalized by the sum of the all the eigenvalues.

For WANG dataset, we also show the result of $m_1 = m_2 = 2, m_3=4$
and $m_1 = m_2 = m_3 = 3$ for 80\% pixel scramble in the last row of
the top part of Figure ~\ref{fig:wang}. For 101 dataset, we add
results $m_1 = m_2 = m_3 = 30$ and $m_1 = m_2 = m_3 = 80$ int the
last row of Figure \ref{fig:101}.

\section{Datasets}

The first benchmark is face databases AT\&T \cite{ATT:Data} in which
there are ten different images of each of 40 distinct subjects.
We use the original size of the image. All 400 images form a
$112\times 92 \times 400$ tensor.

The second image dataset is WANG \cite{WangLW01} which contains
10 categories (Africa, Bench, Buildings, Buses, Dinosaurs,
Elephants, Flowers, Houses, Mountains, and Food) and 100 images for
each category. The original size of the image is either $384\times
256$ or $256\times 384$. We select Buildings, Buses, and Food
categories and resize the images into a $100 \times 100$ size. We
also transform all images into 0-255 level gray images.
The selected images
form a $100 \times 100 \times 300$ tensor.

The third dataset is Caltech 101 \cite{Caltech:101} which contains
101 categories. About 40 to 800 images per category. Most categories
have about 50 images. Collected in September 2003 by Li,
Andreetto, and Ranzato.  The size of each image
is roughly $300 \times 200$ pixels. We randomly pickup 200 images,
resize and transform them into $100\times 100$
0-255 level gray images to form a $100 \times 100 \times 200$
tensor.

\section{Image randomization}

Three types randomization are considered:
block scramble, pixel scramble and occlusion.
In block scramble, an image is divided into n = 2, 4, 8
blocks; blocks are scrambled to form new images
(see Figure ~\ref{fig:atnt},
\ref{fig:wang} and \ref{fig:101}).

In pixel sample, we randomly pick up $\alpha = 40\%, 60\%,80\%$
of the pixels in the image, and randomly scramble them
to form a new image (see Figure ~\ref{fig:atnt}, \ref{fig:wang}, and
\ref{fig:101}).

We also experimented with occulsions with sizes upto half of the
images. We found that occulsion consistently produce smaller
randomization affects and HOSVD results converge to the
unique solution. For this reason and the space limitation,
we do not show the results here.

\section{Main Results}

From results shown in Figures ~\ref{fig:atnt}, \ref{fig:wang}, and
\ref{fig:101}.
we observe the following:
\begin{enumerate}

\item
For all tested real-life data, ParaFac
solutions are not unique, i.e., the converged solution
depends on initial starts.
This is consistent with the non-convex optimization as explained
in \S 2.2.

\item
For all tested real-life data, HOSVD
solutions are unique, although theoretically, this is
not guarrentteed since the optimization of HOSVD is non-convex
as explained in \S 2.1;

\item
For even heavily rescrambled (randomized) real-life data, HOSVD
solutions are also unique;
This is surprsing, given that the HOSVD optimization
are non-convex.

\item
For very severelly rescrambled real-life data and
pure randomly generated data, HOSVD
solutions are not unique.

\item
The HOSVD solution for a given dataset may be
unique for some parameter setting
but non-unique for some other parameter setting.

\item
Whether the HOSVD solution for a given dataset will
be unique can largely be predicted by
inspecting the eigenvalue distribution of the
matrices $F,G,H$.
See next section.

\end{enumerate}

\section{Eigenvalue-base uniqueness prediction}

We found Empirically that the eigenvalue distribution help
to predict whether the HOSVD solution
on a dataset with a parameter setting
is unique or not.

For example, in AT\&T dataset  HOSVD converges in all parameter
settings except in $8\times8$ block scramble with $ m_1 = m_2 = m_3 =
5$. This is because the ignored 3 eigenmodes
have very similar eigenvalues as the first five eigenvalues.
It is ambiguous for
HOSVD to select which of the 8 significant eigenmodes.
Thus HOSVD fails to converge to a unique solution.

But when we increase $m_1, m_2, m_3$ to 10, all 8 significant
eigenmodes can be selected and
HOSVD converges to a unique solution.
This also happens in the other two datasets (see the forth rows in
top part of Figures \ref{fig:wang} and \ref{fig:101}.

For 80\% pixel scramble in dataset WANG, when $m_1 = m_2= m_3 = 5,
10$, HOSVD is ambiguous as to  select eigenmodes because
there are a large number of them with nearly identical eigenvalues
around the cutoff.
However,
if we reduce the dimensions to $m_1 = m_2 = 2, m_3 = 4$ or $m_1 =
m_2 = m_3 = 3$, this ambiguity is gone:
HOSVD clearly selects the top 2 or 3 eigenmodes.
converges (see the last row of the top panel in Figure~\ref{fig:wang}).
This same observation also applies to
Caltech 101 dataset at
80\% pixel scramble in 101
(see the last row of the top part of
Figure~\ref{fig:101}).

\begin{figure}
\center
\includegraphics[scale=0.58]{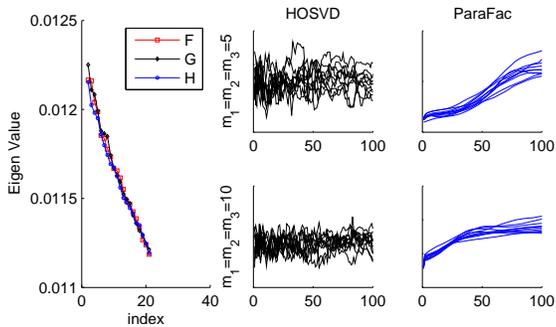}\\
 \caption{HOSVD and ParaFace Convergence on a $100\times 100 \times 100$ random tensor.} \label{fig:random}
\end{figure}

\begin{figure*}
\center
\includegraphics[scale=0.55]{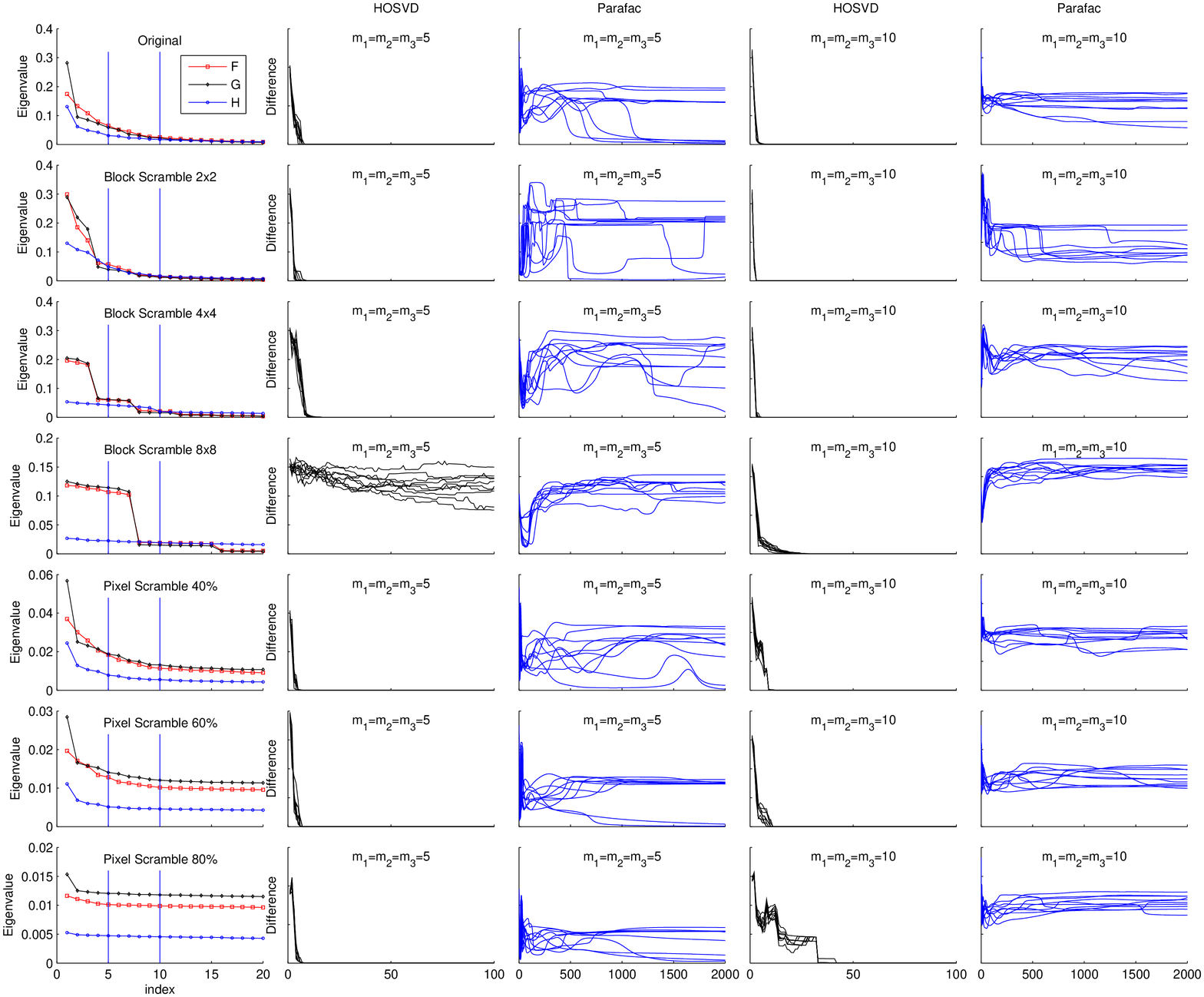}\\
\includegraphics[scale=0.75]{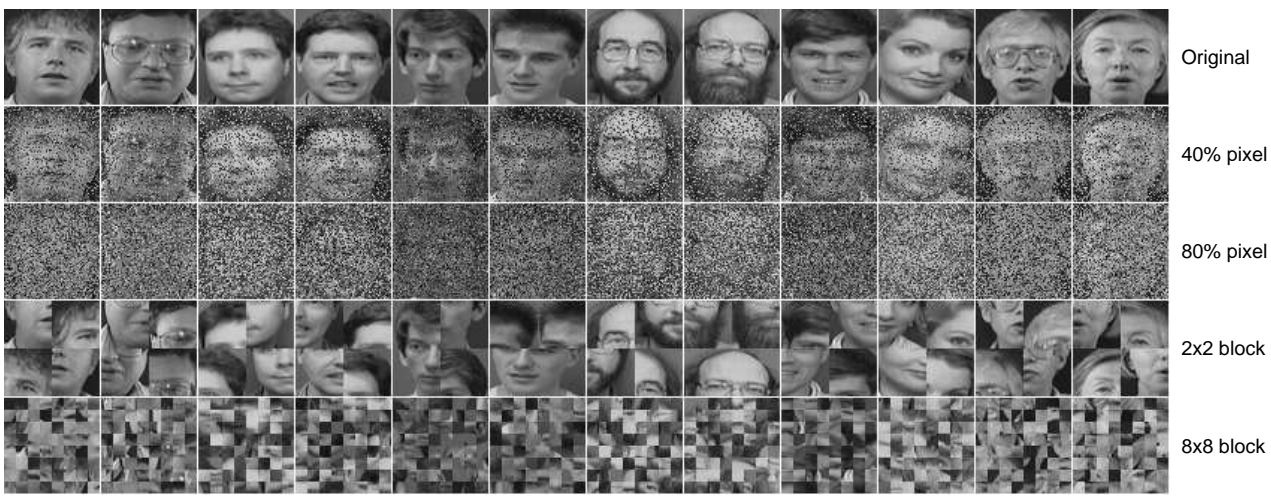}
 \caption{
AT\&T dataset (400 images of size  $112\times 92$ each).
Shown are eigenvalues of $F,G,H$, and
solution uniqueness of HOSVD and ParaFac.}
\label{fig:atnt}
\end{figure*}

\begin{figure*}
\center
\includegraphics[scale=0.55]{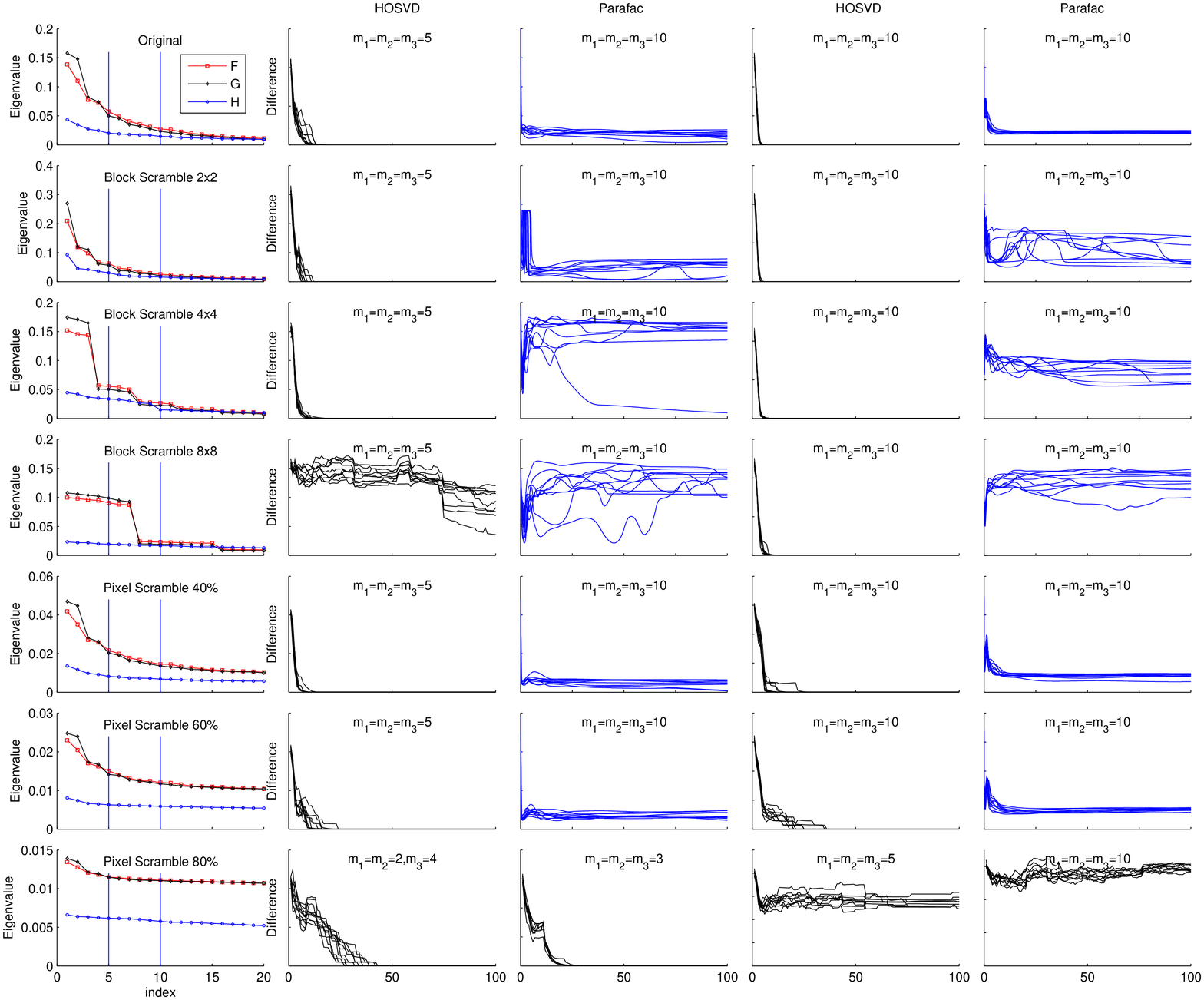}\\
\includegraphics[scale=0.75]{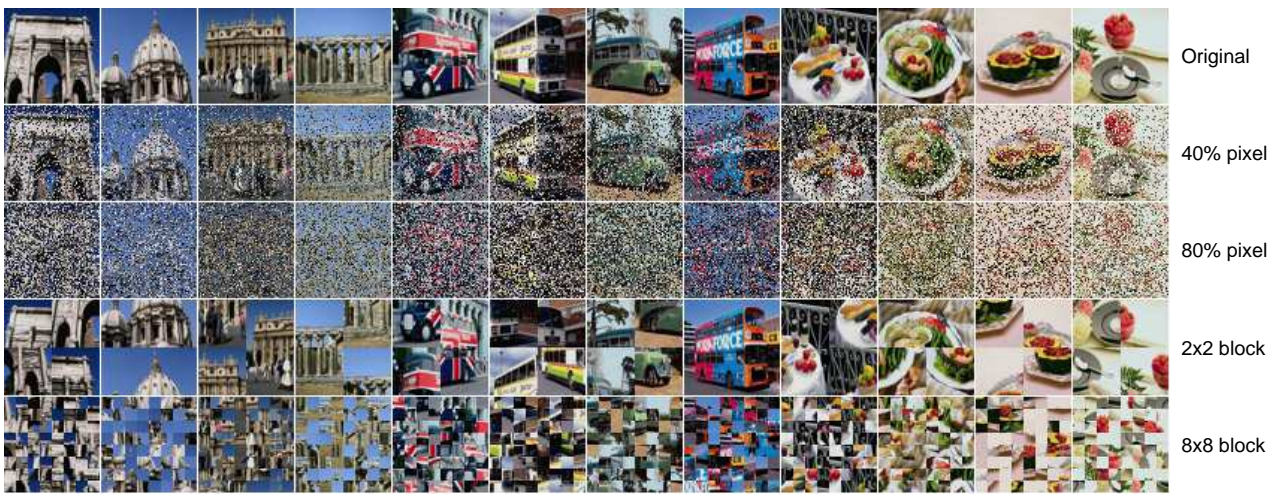}
 \caption{
WANG dataset (300 images, $100\times 100$ size for each).
Shown are eigenvalues of $F,G,H$, and
solution uniqueness of HOSVD and ParaFac.}
\label{fig:wang}
\end{figure*}

\begin{figure*}

\center
\includegraphics[scale=0.55]{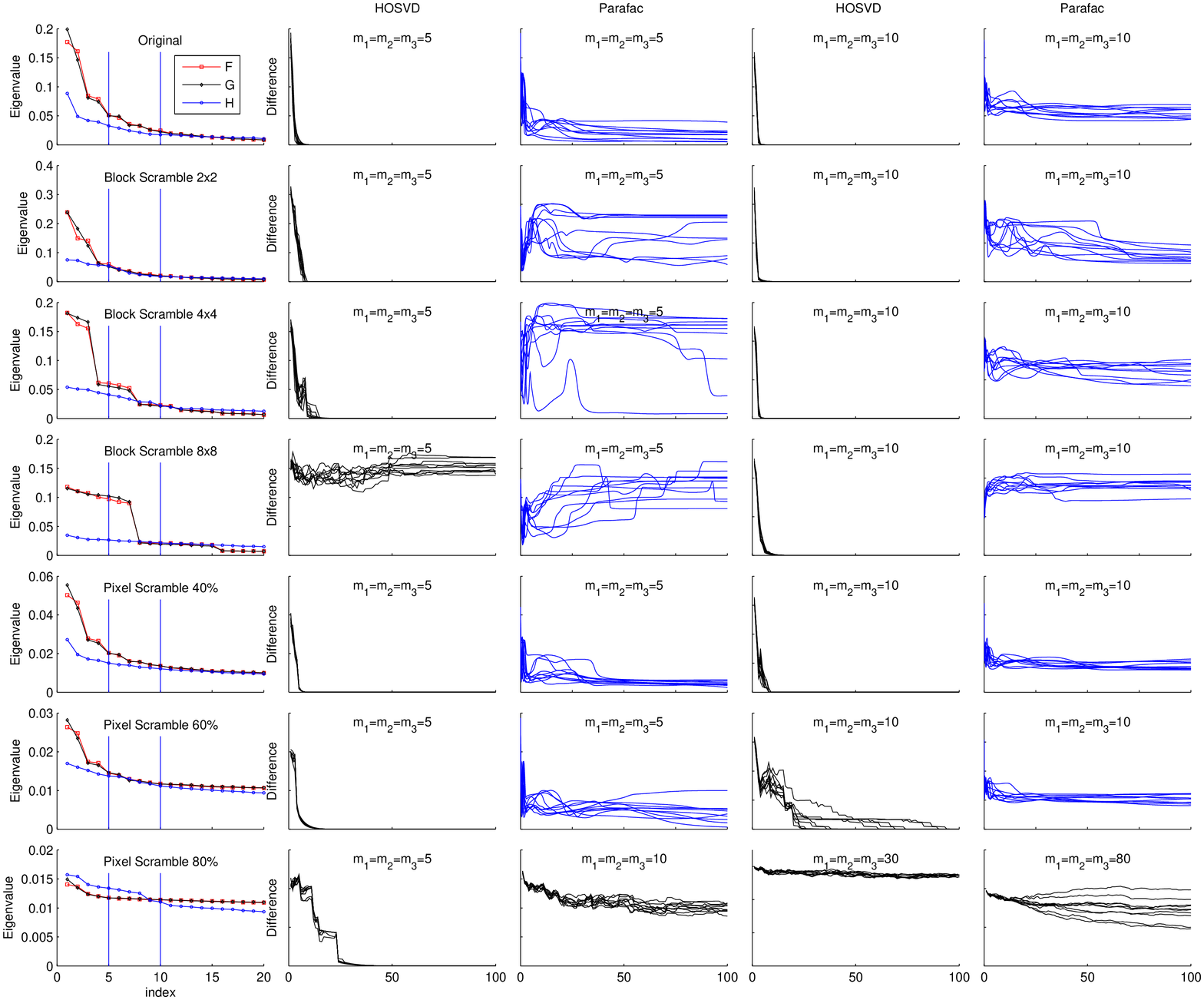}\\
\includegraphics[scale=0.75]{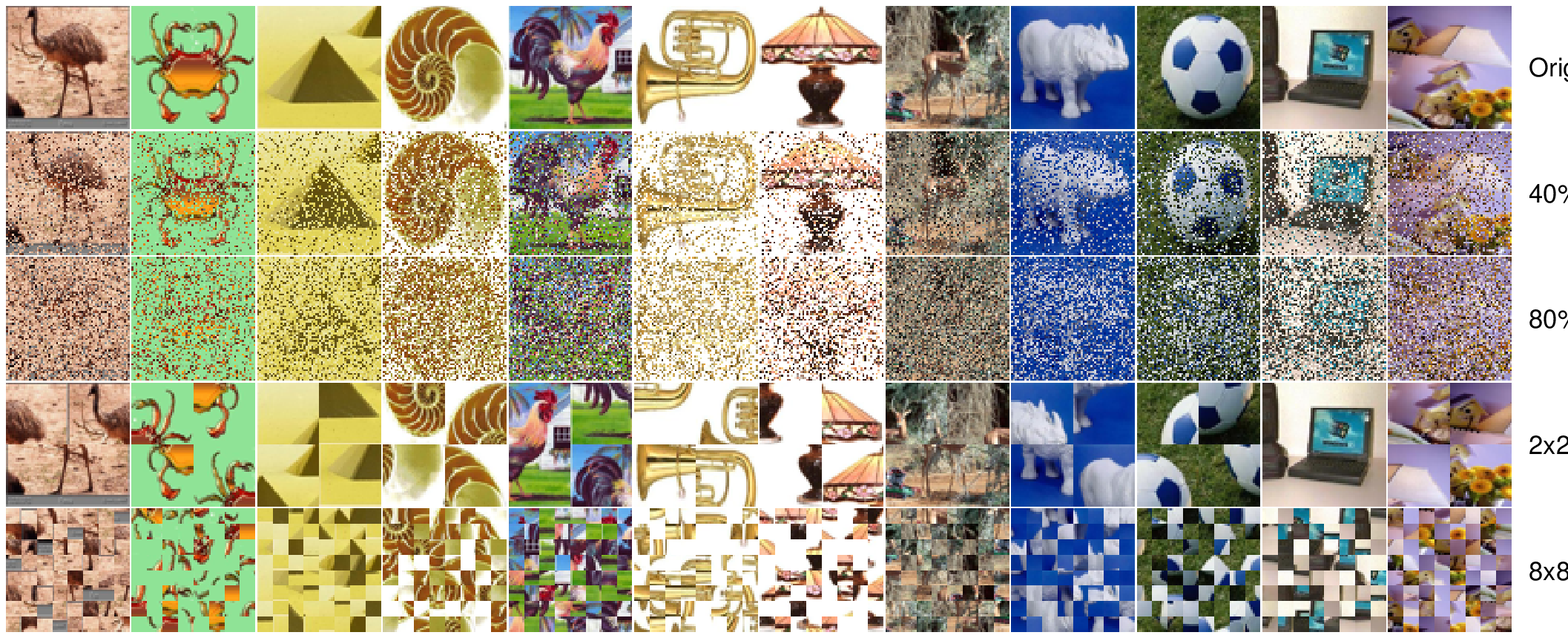}
 \caption{
Caltech 101 dataset (200 images of size $100\times 100$ each).
Shown are eigenvalues of $F,G,H$, and
solution uniqueness of HOSVD and ParaFac.}
\label{fig:101}
\end{figure*}

For random tensor shown
in Figure ~\ref{fig:random}, the eigenvalues are nearly identical
to each other. Thus for both parameter setting ($m_1 = m_2
= m_3 = 5$ and $m_1 = m_2 = m_3 = 10$), HOSVD is ambiguous to
selection eigenmodes and thus does not converge.

We have also investigated the
solution uniqueness problem of
the GLRAM tensor decomposition.
The results are very close to HOSVD.
We skip it
due to space limitation.

\section{Summary}

In summary, for all real life datasets we tested,
the HOSVD solution are unique (i.e., different
initial starts always converge to an unique global solution);
while
the ParaFac solution are almost always not unique.
These finding are new (to the best of our knowledge).
They also surprising and comforting. We can be assured that
in most applications using HOSVD, the solutions are
unique --- the results are reliable and repeatable.
In the rare cases where the data are highly irregular or
severely distored/randomized, our results indicate that
we can predict whether HOSVD solution is unique by
inspecting the eigenvalue distributions.


\end{document}